\documentclass[runningheads]{llncs}
\usepackage[T1]{fontenc}
\usepackage{subcaption}
\usepackage{amsmath}
\usepackage{amssymb}
\usepackage{pifont}
\usepackage{algorithm}
\usepackage{algpseudocode}
\usepackage{algorithmicx}
\usepackage{multirow}
\usepackage{booktabs}
\usepackage{graphicx}
\usepackage{lscape}
\usepackage{enumitem}
\usepackage{wrapfig}

\usepackage{color,soul}
\newcommand\blfootnote[1]{%
  \begingroup
  \renewcommand\thefootnote{}\footnote{#1}%
  \addtocounter{footnote}{-1}%
  \endgroup
}
\usepackage{hyperref}
\usepackage{color}

\begin{document}
\title{WaterMAS: Sharpness-Aware Maximization for Neural Network Watermarking}
\titlerunning{WaterMAS}
\author{Carl De Sousa Trias\inst{1} \and
Mihai Mitrea\inst{1}\and
Attilio Fiandrotti\inst{2,3} \and Marco Cagnazzo\inst{3,4} \and Sumanta Chaudhuri \inst{3} \and Enzo Tartaglione \inst{3}}
\authorrunning{C. De Sousa Trias et al.}
\institute{Telecom SudParis, Institut Polytechnique de Paris, Palaiseau, France  \and
Università degli Studi di Torino, Turin, Italy
\and
LTCI, T\'el\'ecom Paris, Institut Polytechnique de Paris, Palaiseau, France
 \and
Università degli Studi di Padova, Padua, Italy}
\maketitle              % typeset the header of the contribution
\begin{abstract}
Nowadays, deep neural networks are used for solving complex tasks in several critical applications and protecting both their integrity and intellectual property rights (IPR) has become of utmost importance.
To this end, we advance WaterMAS, a substitutive, white-box neural network watermarking method that improves the trade-off among robustness, imperceptibility, and computational complexity, while making provisions for increased data payload and security. WasterMAS insertion keeps unchanged the watermarked weights while sharpening their underlying gradient space. The robustness is thus ensured by limiting the attack's strength: even small alterations of the watermarked weights would impact the model's performance. The imperceptibility is ensured by inserting the watermark during the training process. The relationship among the WaterMAS data payload, imperceptibility, and robustness properties is discussed. The secret key is represented by the positions of the weights conveying the watermark, randomly chosen through multiple layers of the model. The security is evaluated by investigating the case in which an attacker would intercept the key. The experimental validations consider 5 models and 2 tasks (VGG16, ResNet18, MobileNetV3, SwinT for CIFAR10 image classification, and DeepLabV3 for Cityscapes image segmentation) as well as 4 types of attacks (Gaussian noise addition, pruning, fine-tuning, and quantization). The code will be released open-source upon acceptance of the article.\blfootnote{This paper has been accepted for publication at the 27th International Conference on Pattern Recognition (ICPR 2024).}

\end{abstract}
\section{Introduction}
\label{sec:intro}
The deployment of deep neural networks became massive for both industrial and end-user-oriented applications. Such tasks are instantiated in a wide variety of application domains including but not restricted to image/video classification~\cite{krizhevsky2009learning,Lecun1998grad}, object detection~\cite{Yang2018object}, speech recognition and synthesis~\cite{ning2019review}, and audio-visual content compression~\cite{lu2019dvc}. Furthermore, deep neural networks can also serve the purposes of critical tasks, such as autonomous driving for unmanned vehicles~\cite{chng2021roneld}. 
Coming across with the effort to make such models more and more efficient in their tasks, the interest in protecting their intellectual property rights (IPR) and in verifying their integrity emerged some 7 years ago. On the one hand, these models are costly in terms of human skills and computing resources, and protecting their intellectual rights is not only an ethical issue but also an economic one. On the other hand, such deep models can voluntarily or involuntarily be corrupted, resulting in the malfunctioning of the system. %costly damages and, worst case, even in the loss of lives. 
Watermarking can be a solution in both cases. 

Inherited from the multimedia realm~\cite{cox2007digital}, watermarking regroups a family of methodological and applicative tools allowing for imperceptible and persistent insertion of some metadata (watermark) into original content, according to a secret key and under some prescribed security performances. The subsequent watermark detection can serve various purposes among which the most relevant for our study is IPR management, understood as the possibility of unambiguously identifying semantically similar yet digitally different contents, like those obtained after compressing or cropping a video sequence, for instance. The main properties of watermarking are the data payload, the imperceptibility, and the robustness. The \emph{data payload} represents the quantity of information (in bits) that can be inserted and detected for serving the targeted applicative scope (copyright and/or integrity certification). The \emph{imperceptibility} refers to the preservation of the quality of the original content. The \emph{robustness} refers to the property of recovering the mark even when the protected content was subjected to malicious or mundane operations (commonly referred to as attacks). The security property relates to the watermarking system behavior when some partial or total information about the key is available to the attacker. 

Watermarking solutions can also be designed and deployed for deep neural networks~\cite{uchida2017embedding,adi2018turning,li2021survey}. To this end, the generic watermarking properties are reconsidered and extended. First, the data payload concept is kept unchanged, yet the practice results in one-bit solutions (a binary, detected/undetected decision is generally made). Secondly, the imperceptibility property is evaluated by comparing the applicative performances of the watermarked model to the ones provided by a watermarked model trained in similar conditions. Thirdly, the robustness is assessed against transformations involved in the neural network life-cycle, like pruning or quantization, for instance. Finally, the watermark could be either retrieved from the parameters of the model (\emph{white-box} methods~\cite{uchida2017embedding,chen2019deepmarks,kakikura2022collusion,tartaglione2021delving}) or from its inference (\emph{black-box} methods~\cite{adi2018turning,zhang2018protecting}). The neural network watermarking method security is not specifically studied and by default considered to be linked to the key size (as an attempt to avoid the brute force attack) and to its impossibility to be known by the attacker.

In this study, we introduce a new neural network watermarking method that belongs to the white-box category. 
On the one hand, from the conceptual point of view, we establish synergies between the very concept of neural network optimization and neural network watermarking. From the neural network optimization point of view, we use the high dimensionality of models to lock a subset of weights without impacting the final performance of the model. To this end, the findings in~\cite{foret2021sharpnessaware} are leveraged for watermarking purposes, where the sharpness of the loss landscape for the watermarked weights should be maximized. This way, the watermark can be inserted through the entire model, without exploiting any of its peculiarities (architecture, activation function, regularization term, etc.). By considering now that deep neural networks are typically over-parameterized, the watermark size can be thus significantly extended, virtually till the limit set by the model redundancy~\cite{lin2017does,frankle2018lottery,louizos2018learning}.
On the other hand, from the neural network watermarking point of view, we insert a watermark (represented as an image) in a subset of weights. The robustness property is obtained by the inference sensitivity on the watermarked weights: the force of the attacks that can be applied on the watermarked weights is implicitly reduced by the imperceptibility property. The same sensitivity also makes the method reach extreme security: even assuming the attacked identifies the key, they cannot modify/erase the watermark without destroying the model inference quality (this \emph{a priori} consideration being experimentally proved in Table~\ref{tab:keysteal}).

The main contributions of this study can be summarized as follows.
\begin{enumerate}
    \item Starting from the SAM (Sharpness-Aware Minimization~\cite{foret2021sharpnessaware}) setup, where the loss landscape is enforced being maximally flat, we state and solve the inverse problem (Sharpness-Aware Maximization - further referred to as MAS) (Sec.~\ref{sec:SAO}) where the loss landscape can be made steep. 
    \item We define a new neural network watermarking method, referred to as WaterMAS, that leverages the MAS principle for turning the produced inference intrinsically sensitive to the watermarked weights (Sec.~\ref{sec:waterMAS}). 
    \item We carry out a comprehensive set of experiments on conventional neural network watermarking properties (imperceptibility, robustness, and computational complexity) as well as a discussion on data payload and method security (Sec.~\ref{sec:experiments}). 
\end{enumerate}

\begin{figure}[t]
    \centering
    %\advance\leftskip-0.3cm
    \includegraphics[width=\textwidth]{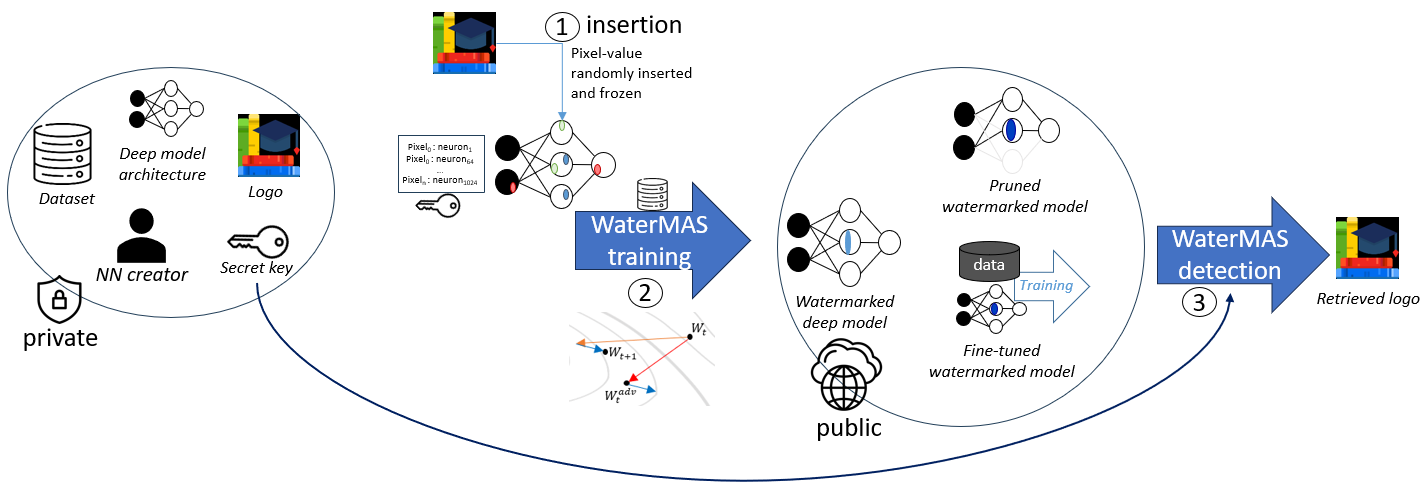}
    \caption{Usage of WaterMAS for sharing a watermarked neural network model, while tracking its usage.}\label{fig:intro}
\end{figure}

The paper is structured as follows. Sec.~\ref{sec:sota} starts with an analysis of the basic concepts for neural network watermarking and exemplifies the key white-box neural network watermarking methods: insertion methods, types of inserted information, secret key, functional properties, and different attacks. It follows by introducing the sharpness-aware minimization grounds. The next Sec.~\ref{sec:SAO} presents the first contribution of the paper, namely the definition of the sharpness maximization problem and the deriving of the underlying algorithm solving this problem. Sec.~\ref{sec:waterMAS} presents a new neural network watermarking method based on the findings in Sec.~\ref{sec:SAO}, as illustrated in Fig.~\ref{fig:intro}. The experimental Sec.~\ref{sec:experiments} starts by imperceptibility evaluation against different methods of the literature, on different tasks and architectures, then assesses the robustness, and lastly, explores the size of information that can be inserted in a specific model and its relationship with security aspects. Finally, Sec.~\ref{sec:conclu} concludes this work and opens perspectives for future works.
\section{Related Works}
\label{sec:sota}
Watermarking tools allow imperceptible and persistent insertion of some metadata into original content. The neural network watermarking field emerged in 2017 with the work of Uchida~et~al.~\cite{uchida2017embedding}, followed by Adi~et~al.~\cite{adi2018turning} and Zhang~et~al.~\cite{zhang2018protecting} in 2018, thus establishing the earliest taxonomy: white-box \textit{vs} black-box watermarking. In the white-box case, the watermark is retrieved from the parameters of the model while the black-box scenario corresponds to the situation where the watermark is retrieved from the inferences of the model. 

\subsection{Neural network watermarking} Uchida~et~al.~\cite{uchida2017embedding} is a white-box watermarking method inserting the watermark into an arbitrarily selected layer, by using a regularization term that projects the parameter on a space (the secret key) a binary watermark. The regularization term is obtained by computing the binary cross-entropy between the projected weights on the secret space and the targeted watermark. At the detection phase, the watermarked layer is projected on the secret space to obtain the watermark. To meet the imperceptibility criterion, the regularization term is weightily added to the loss of the original task. The robustness is expected to be met by the design of the term which spreads the information of the watermark within all the weights of the layer. In other words, this method implicitly considers that thanks to the regularization term, no matter how a modification is applied to the watermarked layer, the watermark will be detected as long as the inference is not severely downgraded.  
The imperceptibility is evaluated by comparing the final accuracy of un-watermarked and watermarked models. The robustness is assessed against fine-tuning (performed as additional training for half the number of embedding/training epochs) and magnitude pruning (setting to zero up to $95\%$ of the layer). 
Methods with similar principles to \cite{uchida2017embedding} can be found in \cite{chen2019deepmarks,Li2021spread}. For instance, in \cite{chen2019deepmarks} the projection is done on the output of the layer instead of its weights. In \cite{Li2021spread}, spread-transform dither modulation is considered as the insertion side.

Tartaglione~et~al.~\cite{tartaglione2021delving} follows a different approach. It no longer considers a specific layer but randomly selects a set of parameters to be watermarked. For the insertion procedure, the pixel values of an image (watermark) are inserted and frozen in randomly selected weights: the correspondences among the pixels in the watermark and the locations of the parameters are kept secret (and represent the key). The design of this method is meant to ensure sensitive watermarked weights: their small variation will impact the performance of the model. During training, at each step, $R$ replicas are created by adding noise to the watermarked weights, and the loss of the $R$ replicas is maximized, thus acting as a regularization term. Hence, such regularization terms can be weightily added to the loss of the original task to meet the imperceptibility criteria. The robustness criterion is ensured by the replicas which randomly explore the loss landscape around the watermarked weights and maximize it. During training, the watermarked parameters are frozen and the original cost function is computed.  At the detection phase, the watermarked parameters are retrieved by looking at their location. Note that the robustness of \cite{tartaglione2021delving} differs from the one in \cite{uchida2017embedding} in its very nature: this time, the potential force of the modifications of the watermarked weights is restricted, as it would implicitly downgrade the inference performance. 
The imperceptibility is evaluated by comparing the final accuracy of un-watermarked and watermarked models, while the robustness is assessed against fine-tuning (performed as additional training for half the number of embedding/training epochs) and quantization (reducing the bits representation of the weights).

Li~et~al.\cite{li2022fused} embeds a binary watermark in the sign of the most significant weights of a model. For the insertion procedure, the binary watermark of length $m$ is mapped to $\{-1,1\}_m$ and modulated by a pseudo-random generator. The selected neurons are chosen according to three pruning methods namely: network slimming, efficient filter, and entropy. In detail, the network slimming method selects the neurons according to the sum of their absolute parameters value, the efficient filter method selects the neurons according to the magnitude of the scale in the batch norm layer, and the entropy method selects the neurons according to the entropy value of their output for a subset of inputs. Hence, each method selects $r$ neurons, then the final list of selected neurons is obtained by intersecting the three obtained lists, finally, $m$ weights are randomly selected to carry the watermark, and their position is kept secret (and represents the key).  The mark is inserted in the sign of the parameters with the highest $\mathcal{L}_1$-norm of the selected neurons. After inserting the watermark the model is fine-tuned for an additional 10 epochs without constraint on the training. 
At the detection phase, the selected parameters are retrieved and their sign is recovered, modulation is reversed to obtain the watermark.
To evaluate imperceptibility the final accuracy of watermarked and unwatermarked models is compared while the robustness is evaluated against fine-tuning, pruning, and watermark overwriting.  

Lv~et~al.~\cite{lv23watermarking} embeds the weight of the encoder of an autoencoder model (HufuNet) in the watermarked model. First, the HufuNet model is trained to reconstruct images of a given dataset. For the insertion procedure, the position of the watermarked weights is obtained by computing the hash function over the Decoder (secret key) while the parameter values of the encoder are inserted. During training, the model is trained according to a specific loss that constrains the evolution of the parameter. Since the watermarked weights might evolve, in each $n$ epoch the watermark parameters are retrieved and the retrieved HufuNet model is fine-tuned with a frozen decoder until reaching back performance, the watermarked weights are inserted again, and the training of the watermarked model resumed. At the detection phase, the watermark parameters are extracted using the Decoder, and the reconstructed HufuNet is evaluated on its tasks. The imperceptibility criteria is evaluated by comparing unwatermarked and watermarked models while the robustness is evaluated against fine-tuning, transfer learning, and pruning.
\subsection{Sharpness-aware minimization}
 Model generalization in deep learning is a critical area of research, recurrently addressed by the Deep Learning community~\cite{caruana2000overfitting,neyshabur2017exploring}. Various advanced techniques have been created, focusing either on adjusting the model itself~\cite{ioffe2015batch} or on enhancing the dataset through augmentation~\cite{Jiang2020Fantastic,zhang2018mixup}. The connection between flat regions in the loss landscape and the underlying model generalization capability was brought forth back already in 1995~\cite{hochreiter1995flat} and it has been empirically studied afterward~\cite{KeskarMNST16,neyshabur2017exploring}. Subsequently, multiple approaches leading to maximally flat regions have been identified, in the seek for better generalization ~\cite{ChaudhariCSL16,Mobahi16,foret2021sharpnessaware}.

In the last few years, Sharpness-Aware Minimization (SAM) was advanced with the goal of being both efficient and effective in enforcing flatness for the achieved solutions, employing a local linear approximation for the loss~\cite{foret2021sharpnessaware}. Starting from this, other variants of SAM have been proposed, achieving outstanding results in several demanding tasks which include continual learning~\cite{deng2021flattening} and federated learning~\cite{caldarola2022improving,dai2023fedgamma}. 

\begin{wraptable}{r}{0.5\textwidth}
\vspace{-20pt}
\caption{Table of notations.} \label{tab:notation}
\vspace{-5pt}
\resizebox{.5\textwidth}{!}{
\begin{tabular}{ c l }
 \toprule
%  Symbol  & Definition \\
$X \in \mathbb{R}^N$ & the watermark\\
$x_i$ & a single element of the watermark \\
$W$ & The parameters of the model\\
$W_X \subset W$ & a subset of the parameters\\ 
$\overline{W}_X$ & the complementary subset\\
$\mathcal{L}$  & the loss of the model \\
$\nabla$ & the gradient\\
$\eta$ & the learning rate\\ 
\bottomrule
\end{tabular}
}
%\vspace{-20pt}
\end{wraptable}
In our work, we are building on top of~\cite{foret2021sharpnessaware} for considering the shape of the loss landscape in an efficient way. Specifically, we like to ensure \emph{sharpness} for a subset of parameters where we embed our watermark. Unlike approaches like~\cite{tartaglione2021delving} that rely on memory and computation-wise expensive sampling process, we leverage the locally linear approximation of the loss to find the ``worst-case'' direction of the gradient in the watermarked parameters space.
\section{Sharpness-aware optimization}
\label{sec:SAO}
This section starts by reconsidering the SAM algorithm presented in \cite{foret2021sharpnessaware} and designed to flatten the loss landscape of all the weights of the model for increasing the model generalization. Then, our algorithm for Sharpness-aware Maximization \textbf{MAS} is introduced: it aims at sharpening the loss landscape of a subset of weights which will serve in the next section \ref{sec:waterMAS} to carry the watermark. Notations are introduced in Table~\ref{tab:notation}.

\subsection{Sharpness-aware minimization (SAM)}
\label{sec::SAM}
\begin{wrapfigure}{r}{0.4\textwidth}
    \centering
    \vspace{-22pt}
    %\advance\leftskip-0.3cm
    \includegraphics[width=0.4\textwidth]{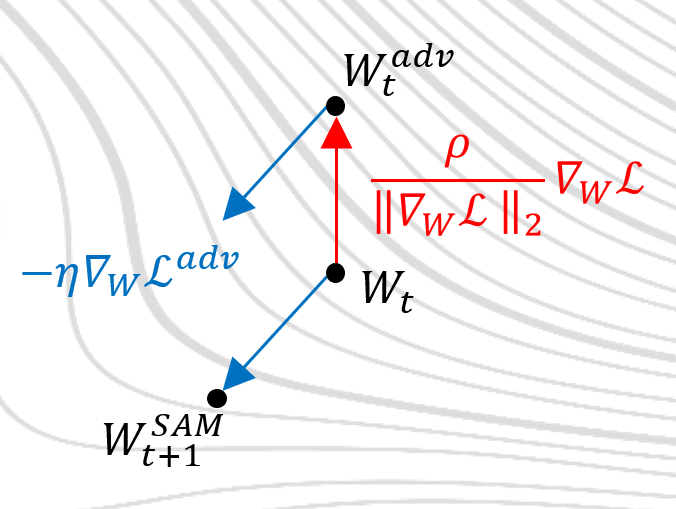}
    \caption{Schematic of the parameter update in SAM method.}
    \label{fig:SAM}
    %%%\vspace{-15pt}
\end{wrapfigure}
In \cite{foret2021sharpnessaware}, a sharpness-aware minimization optimization algorithm (SAM) was advanced to achieve model generalization by using only two gradient computations per iteration, as synoptically presented in Figure~\ref{fig:SAM}. The first gradient is computed from all the parameters of the model ($\nabla_{W}\mathcal{L}$). Then, $W_t^{adv}$ is obtained by maximizing $\mathcal{L}$, i.e. by following the gradient $\nabla_{W}\mathcal{L}$ properly scaled using $\rho$:
\begin{equation}
    \label{eq:gadvSAM}
    W^{adv}_t = W_t + \frac{\rho}{\|\nabla_{W}\mathcal{L}\|_2}\nabla_{W}\mathcal{L}.
\end{equation}
This way, $W_t$ is projected to the point, on the hypersphere of radius $\rho$ centered to $W_t$, where the loss is maximal (under the assumption the loss is locally linear, meaning that this assumption is valid for low values of $\rho$). The objective of SAM is to ensure the local flatness of the loss: by minimizing the loss at ${W}^{adv}$, this can be successfully achieved. This gradient is the projected to $W_t$:
\begin{equation}
    \label{eq:gasm}
    W_{t+1}^{SAM} = W_t - \eta\nabla_{W}\mathcal{L}^{adv}.
\end{equation}

Empirical results on multiple common architectures, including ResNets and state-of-the-art image classification datasets, validate the approach and quantitative analyses demonstrate an improved achieved flatness. In the next subsection, we will build on top of this idea to enforce sharpness on a subset of parameters.

\subsection{Sharpness-aware maximization (MAS)}
\label{sec::MAS}
\begin{figure}[t]
\centering
\begin{subfigure}[h]{0.6\textwidth}
    \includegraphics[width=\textwidth]{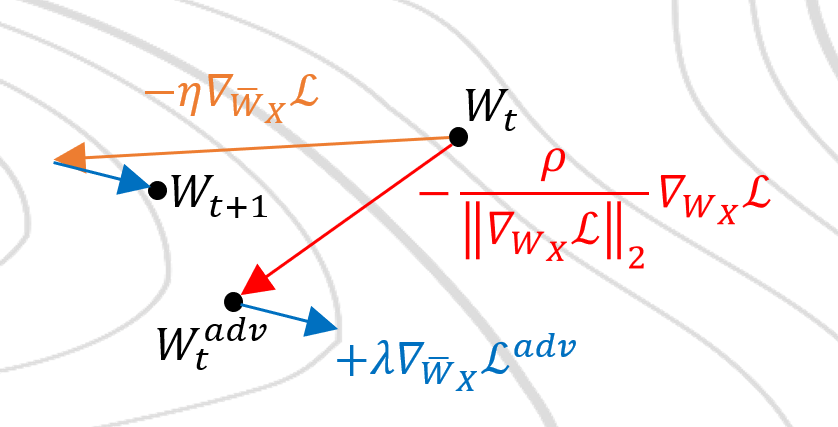}
    \caption{Schematic of the parameter update in MAS.}
    \label{fig:overviewmas}
\end{subfigure}
\begin{subfigure}[h]{0.49\textwidth}
    \includegraphics[width=\textwidth]{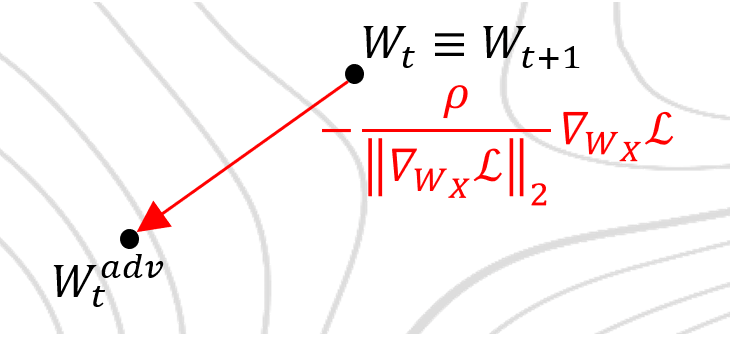}
    \caption{Projection in the space $W_X$.}
    \label{fig:projw}
\end{subfigure}
\begin{subfigure}[h]{0.49\textwidth}
    \includegraphics[width=\textwidth]{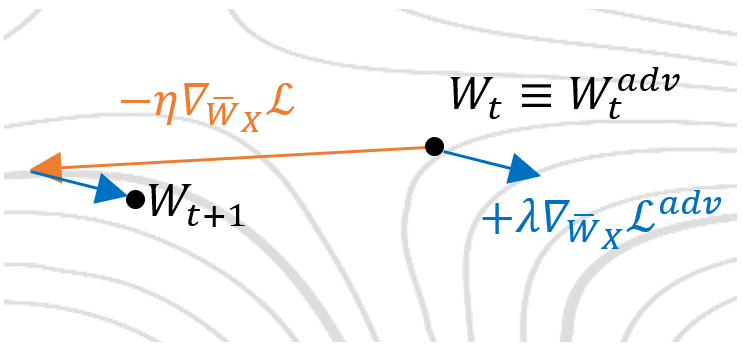}
    \caption{Projection in the space $\overline{W}_X$.}
    \label{fig:projuw}
\end{subfigure}
\caption{Schematic of the parameter update in MAS method for the different loss landscape (a) $W$  (b) $W_X$ and (c) $\overline{W}_X$.}
\label{fig:MAS}
\end{figure}
Let us split the set of parameters $W$ belonging to our model in two distinct subsets of weights: $W_X$, consisting of some frozen parameters to which we want to enforce sharpness for a target loss $\mathcal{L}$ (while still minimizing it), and $~{\overline{W}_X=W\setminus W_X}$. The choice of the loss function depends on the specific problem the model is designed to address, \textit{eg.} cross-entropy for classification models.
The objective is thus to drive the subset of weights $W_X$ (by properly modifying $\overline{W}_X$ only) to a sharp region. We follow a three-step computation strategy, which is summarized in Fig.~\ref{fig:overviewmas}. 

First, we solely minimize the loss $\mathcal{L}$ wrt. $W_X$, by projecting $W_t$ to an hypersphere of radius $\rho$ limitedly to the subspace $W_X$, thus obtaining $W^{adv}_t$:
\begin{equation}
    \label{eq:gadv}
    W^{adv}_t = W_t - \frac{\rho}{||\nabla_{W_X}\mathcal{L}||_2}\nabla_{W_X}\mathcal{L}.
\end{equation}
This projection is represented in Fig.~\ref{fig:projw}. Note that in the space $\overline{W}_X$, the projections of $W_t$ and of $W^{adv}_t$ are at the same point. 

Then, we aim at maximizing the loss on $W^{adv}_t$. This way, we are enforcing that, at a given distance $\rho$, the loss increases when moving in the $W_X$ space.

Finally, we join the adversarial loss maximization term computed in the previous step with the traditional loss minimization on $\overline{W}_X$ (as shown in Fig.~\ref{fig:projuw}):
\begin{equation}
    \label{eq:gmas}
    W_{t+1} = W_t - \eta\nabla_{\overline{W}_X}\mathcal{L} + \lambda \nabla_{\overline{W}_X}\mathcal{L}^{adv}.
\end{equation}
In conclusion, the model will minimize the loss of the given task while aiming for a sharpened region of the $W_X$' loss landscape. In the next section, \eqref{eq:gmas} will be leveraged for watermark insertion: the non-watermarked weights can be updated while minimizing the loss function.

\section{WaterMAS}
\label{sec:waterMAS}
This section advances a new neural network watermarking method \textbf{WaterMAS}. It consists of three steps: (i) selecting the parameters of the model that will carry the watermark (note that unlike most of the white-box neural network watermarking methods, in WaterMAS, the watermarked weights can be located in the whole model rather than in a specific layer); (ii) inserting the watermark in those parameters, and (iii) ensuring the trade-off between robustness and imperceptibility through training. This section is organized as follows. First, information related to data payload, secret information, and weight selection is presented in Sec.~\ref{sec::select}. Secondly, the very training method is advanced by using principles introduced in Sec.~\ref{sec::MAS}. Finally, the detection issues are discussed in Sec.~\ref{sec::detec}.

\subsection{Data payload, secret information, and weight selection}
\label{sec::select}
WaterMAS inserts an image of size $Height\times Width\times 3$ (X) obtained by shifting the integer pixels values towards floating point $x_i \in [0, 1]$. The underlying data payload is 1-bit (image inserted or not) but the impact and the possibility of the different image size will be discussed in Sec.~\ref{sec::secu}.
This image is to be inserted in some model parameters that should be selected in a manner that makes them indistinguishable from the other parameters.
%As we are going to see, this apparent restriction will not be an obstacle during the training phase, as state-of-the-art performances will be achieved in any case. It has been shown that typical learning scenarios involve the research of wide minima, whose existence is related to the typical over-parametrization of ANN models~\cite{lin2017does,frankle2018lottery}. It is not a case, in fact, that deep models can be hugely compressed, as most of the parameters are not necessary after the training phase~\cite{tartaglione2018learning,molchanov2017variational,louizos2017learning}. Exploiting this, we are going to look for a particular solution of $W$ resulting in a \emph{narrow} minima in the subspace $W_x$.
To achieve this, two mechanisms will be considered. The first mechanism consists of randomly sampling $W_X$ parameters to carry the watermark throughout the whole model; this way, a random association between the $i$-th element of the watermark $x_i$ and the $j$-th parameter of the neural network $w_j$ is established. This association serves as a secret key.
The second mechanism refers to the way in which these parameters are modified so as to carry the watermark. Assuming, as presented in \cite{glorot10understanding}, that all the parameters in the model are distributed according to a Gaussian distribution with mean $\mu$ and standard deviation $\sigma$, each inserted element of the watermark follows the equation:
\begin{equation}
    w_j = 2\sigma(x_i - 0.5) + \mu\ \; \; \; \forall\ w_j\in W_X.
\end{equation}

\subsection{Training procedure}
\label{sec::trainMAS}
The training procedure is designed so as to (1) keep the watermarked parameters unchanged (so as to preserve the watermark) and (2) to maximize their sharpness (so as to turn the inference very sensitive on those values).
This is achieved by using \textbf{MAS} introduced in Sec.~\ref{sec::MAS} and according to the \textbf{while} loop of Algorithm~\ref{alg:MAS}. The two subsets of weights are obtained after the selection and insertion occurred (lines $1$\&$2$) and each iteration of training consists of four steps.  (i) A first gradient is computed on the model $W_t$. (ii) $W^{adv}_t$ is obtained by updating the watermarked weights $W_X$ in the direction that minimize the loss. (iii) A second gradient is computed on this obtained model $W^{adv}_t$ and maximized. (iv) The second obtained gradient is added using a hyperparameter $\lambda$ (strength of the additional terms) and applied only to the non-watermarked weights to obtain $W_{t+1}$. 
Hence, the watermarked parameters do not change during the training process while aiming to shrink the loss landscapes, making them sensitive in inference.

\begin{algorithm}[t]
\caption{WaterMAS algorithm}\label{alg:MAS}
\begin{algorithmic}[1]
\Require $\mathcal{L}${(the loss function)} $x_i,y_i${(a pair of inputs, labels) } $X${(the watermark)} 
\State Select a subset of weight to carry the watermark $\boldsymbol{W}_{X}$
\State Substitute the value of $\boldsymbol{W}_{X}$ by $X$ using a mapping function $M$ (kept secret)
\While{Not converged}{
\For{$\forall x_i,y_i$}{
\State Compute gradient on the model $\nabla_{W}\mathcal{L}$ \label{line:1}
\State Obtain $W^{adv}_t$ by only updating the watermarked weights using \eqref{eq:gadv}\label{line:2}
\State Compute gradient on the adversarial model$\nabla_{\overline{W}_{X}}\mathcal{L}^{adv}$ \label{line:3}
\State Add the obtained gradient in \ref{line:3} to the original cost function \label{line:4}
\State Update only the non-watermarked weights according  to as in \eqref{eq:gmas}\label{line:5}
\EndFor}
\EndWhile}
\end{algorithmic}
\end{algorithm}

\subsection{Detection procedure}
\label{sec::detec}
The extraction of the watermark is obtained by retrieving the value of the watermarked weights and by inverting the association between those weights and the inserted image. This extracted image can be compared to the inserted image while using qualitative (human decision) or quantitative (Pearson's correlation). In the end, a binary decision is made (that is, whether the recovered and inserted images are identical or not), and the data payload is 1 bit.\\

In this section, we have introduced a new watermarking method that inserts a watermark by substituting the value of the weights according to secret information only dependent on the number of elements in the watermark (Sec.~\ref{sec::select}). During training, the watermark weights are frozen and the \emph{MAS} term is used to enhance the robustness of those watermarked parameters (Sec.~\ref{sec::trainMAS}). In the next section, an experimental study asses the WaterMAS performances.

\section{Experimental results}
\label{sec:experiments}
In this section, WaterMAS performances are experimentally assessed. The number of watermarked weights $W_X$ remains fixed for the imperceptibility and robustness sections, namely $32\times 32 \times 3$.  We empirically set $\rho=10^{-2}$ and $\lambda=10^{-5}$ through a grid-search evaluated on a ResNet-18 trained on CIFAR-10. WaterMAS is benchmarked against four state-of-the-art methods, namely \cite{uchida2017embedding,tartaglione2021delving,li2022fused,lv23watermarking}, with their reference hyperparameters; \cite{uchida2017embedding,tartaglione2021delving,lv23watermarking} codes were available while \cite{li2022fused} code has been implemented for the present study purposes. 

\subsection{Testbed}
\label{sec::Testbed}
The results are performed on $4$ architectures for classification, namely ResNet18~\cite{he2016residual}, VGG16~\cite{simonyan14vgg}, MobileNetV3s~\cite{howard19mobilenet}, and SwinT~\cite{liu21SwinTH} on CIFAR-10~\cite{krizhevsky2009learning}, and on one architecture for image segmentation, namely DeepLabV3~\cite{chen18DeepLab} on Cityscapes~\cite{cordts2016cityscapes}. Note that throughout the experiments, we tried to impose reproducible conditions by setting the seed; yet, the speculative execution of CUDNN often introduces undesirable sources of randomness negligible to the final performance evaluation~\cite{Pytorchrepro}.
For performance evaluation of the task, the complementary accuracy (acc) ($\dag$) defined as $(1-\text{acc})$ is used for the classification tasks, and the complementary mean Intersection over Union (mIoU) ($\ddag$) defined as $(1-\text{mIoU})$ is used for the image segmentation task. For the robustness evaluation, 4 removal attacks are considered.
\begin{enumerate}
    
\item  \textbf{Gaussian noise addition}: a random noise is added to the model to impact the watermark. Our hyperparameter $S\in [1,5]$ corresponds to the ratio between the standard deviation of the added noise to the standard deviation of the aimed layer while both means are equal.

\item  \textbf{Fine-tuning}: the training of the model is resumed for some additional epochs $E \in [1,25]$. 

\item  \textbf{Magnitude pruning}: a portion $P \in [0.1,0.9]$ of neurons as set to zero according to their $L1$-norm. This attack aims to compress the model. 

\item   \textbf{Quantization}: compress the model by reducing the number of bits $~{B \in [2,16]}$ of the floating representation of the parameters. 

\end{enumerate}
Finally, a discussion about the security, data payload, and computational complexity is devoted in the last section presenting a cryptography attack.

%\vspace{+6pt}
\subsection{Imperceptibility evaluation}
\label{sec::imp}
%\vspace{-18pt}
\begin{table*}[t]
\centering
\caption{Imperceptibility evaluation. Each model has been initialized and trained with the same hyperparameters. All experimental values are multiplied by 100.}%The first row corresponds to a model trained without embedding a watermark.} 
\label{tab:imperceptibility}
\resizebox{0.9\textwidth}{!}{
\begin{tabular}{ c c c c c c }
 \toprule
   \multirow{2}{*}{\bf Method}  &  \multicolumn{4}{c}{\textbf{CIFAR10} } &\bf Cityscapes \\
  %  &  \multicolumn{4}{c}{($^\dag$top-1 accuracy error)} & ($^\ddag$complementary mIoU) \\
    &  VGG16 & ResNet18 & MobileNet  & SwinT & DeepLab-v3\\
\midrule
Unwatermarked  & \it 9.56$^\dag$ &  15.69$^\dag$ &  \it 27.11$^\dag$ & \it 16.28$^\dag$  & \it 29.52$^\ddag$ \\
 \midrule
Uchida~et~al.~\cite{uchida2017embedding}  & \it 9.68$^\dag$ &  \it 13.50$^\dag$ & \it 27.16$^\dag$ & \it 21.76$^\dag$ & \it 36.23$^\ddag$ \\
% \midrule
%KAKI*  & \it  9.56$^\dag$ &  \it 13.66$^\dag$  & \it 27.04$^\dag$  & \it 16.42$^\dag$& \it 29.53$^\ddag$   \\
 \midrule
Tartaglione~et~al.~\cite{tartaglione2021delving}   & \it 9.91$^\dag$ &  \it 12.77$^\dag$ & \it 25.27$^\dag$ & \it 18.47$^\dag$ & \it 30.14$^\ddag$ \\
\midrule
Li~et~al.~\cite{li2022fused}   & \it 9.63$^\dag$ &  \it 15.43$^\dag$ & \it 25.38$^\dag$ & \it 21.07$^\dag$ & \it 29.84$^\ddag$  \\ 
\midrule
Lv~et~al.~\cite{lv23watermarking}   & \it 14.80$^\dag$ & \it 19.60$^\dag$ & \it 33.90$^\dag$ & $\star$ & $\star$\\
 \midrule
WaterMAS  & \it 10.64$^\dag$   &  \it 13.58$^\dag$ & \it 26.34$^\dag$ & \it 21.15$^\dag$ & \it 29.90$^\ddag$  \\
\bottomrule
\end{tabular}
}
\end{table*}

Imperceptibility evaluates the impact of the watermark insertion in the inference. In neural network watermarking, some insertion methods act through the training, thus making the creation of the model intrinsically linked to the watermark. Hence, the imperceptibility evaluation is done by comparing the results of watermarked and unwatermarked models. For all the setups, each model has been initialized with the same seed, trained for the same number of epochs, $200$, using an SGD optimizer with $lr=0.1$ (except for VGG16 $lr=0.01$ and SwinT $lr=0.001$), momentum$=0.9$, weight decay=$10-4$, and a scheduler which divide the learning rate by $10$ after epoch $100$ and $150$. The objective was not to obtain state-of-the-art accuracy but rather to show the impact of the different methods with an identical training setup (in an end-user setup, the hyperparameter can be fine-tuned for each setup to increase the performance of the models). The results are displayed in Table~\ref{tab:imperceptibility} and show similar imperceptibility for all the methods. For Uchida's method on two setups, namely SwinT and DeepLab, we can observe that the regularization term has a stronger impact on the performance, either positively or negatively. For Lv's method, two configurations were not implemented (SwinT and DeepLab), as indicated by $\star$. For VGG16, ResNet18, and MobileNet, the watermarking procedure impacts the performance of the model, indicating that this method needs specific tuning of hyperparameters depending on the configuration.
%\vspace{-10pt}
\begin{table*}[h]
    \centering
\caption{Robustness evaluation of WaterMAS against four removal attacks.
%Gaussian noise addition, magnitude pruning, fine-tuning, quantization. 
The performance metric is multiplied by 100.} \label{tab:robustness}
\resizebox{.7\columnwidth}{!}{
\begin{tabular}{c c c c c c c c c}
 \toprule
     & &   \multicolumn{4}{c}{\textbf{CIFAR10} } &\bf Cityscapes \\
   & &  VGG16 & ResNet18 & MobileNet & SwinT  & DeepLab \\
\midrule
 \multirow{5}{*}{\rotatebox[origin=c]{90}{Gaussian}} & $\boldsymbol{S}\!=\!0$ & \it 10.64$^\dag$ & \it 13.58$^\dag$ & \it 26.34$^\dag$ & \it 21.15$^\dag$ & \it 29.9$^\ddag$\\
\cmidrule{2-7}
& $\boldsymbol{S}\!=\!1$ & \it 10.68$^\dag$  & \it 13.66$^\dag$ &  \it 26.823$^\dag$ & \it 21.30$^\dag$ & \it 30.02$^\ddag$\\

\cmidrule{2-7}
 
& $\boldsymbol{S}\!=\!2$ & \it 10.61$^\dag$  & \it 13.56$^\dag$ & \it 27.00$^\dag$ & \it 21.13$^\dag$ & \it 30.30$^\ddag$\\

\cmidrule{2-7}
 
& $\boldsymbol{S}\!=\!5$ & \it 10.92$^\dag$  & \it 23.29$^\dag$ &  \it 27.14$^\dag$ &  \it 21.49$^\dag$ & \it 36.07$^\ddag$\\
\midrule
\multirow{5}{*}{\rotatebox[origin=c]{90}{Pruning}} & $\boldsymbol{P}\!=\!1$ & \it 10.66$^\dag$ & \it 13.65$^\dag$ & \it 26.75$^\dag$  & \it 20.98$^\dag$  & \it 30.02$^\ddag$\\

\cmidrule{2-7}
 
& $\boldsymbol{P}\!=\!2$ & \it 10.76$^\dag$ & \it 14.05$^\dag$ & \it 26.97$^\dag$ & \it 21.73$^\dag$  & \it 65.24$^\ddag$\\

\cmidrule{2-7}
 
&  $\boldsymbol{P}\!=\!5$ & \it 12.08$^\dag$ & \it 14.91$^\dag$ & \it 44.47$^\dag$ & \it 29.98$^\dag$ & \it 97.35$^\ddag$\\

\cmidrule{2-7}
 
&  $\boldsymbol{P}\!=\!9$ & \it 87.80$^\dag$  & \it 78.39$^\dag$ &  \it 89.89$^\dag$ & \it 86.20$^\dag$ &  \it 99.98$^\ddag$\\
\midrule

\multirow{5}{*}{\rotatebox[origin=c]{90}{Fine-tuning}} & $\boldsymbol{E}\!=\!0$ & \it 10.64$^\dag$ & \it 13.58$^\dag$ & \it 26.34$^\dag$ & \it 21.15$^\dag$ & \it 29.9$^\ddag$\\
\cmidrule{2-7}
& $\boldsymbol{E}\!=\!5$ & \it 9.76$^\dag$ &  \it 12.98$^\dag$ & \it 24.88$^\dag$ & \it 20.88$^\dag$ & \it 28.5$^\ddag$\\

\cmidrule{2-7}
 
 & $\boldsymbol{E}\!=\!10$ & \it 9.73$^\dag$ & \it 12.25$^\dag$ & \it 25.38$^\dag$ & \it 20.06$^\dag$ & \it 27.77$^\ddag$\\

\cmidrule{2-7}
 
&  $\boldsymbol{E}\!=\!25$ & \it 9.58$^\dag$ & \it 11.12$^\dag$ & \it 25.43$^\dag$ & \it 20.60$^\dag$ & \it 27.21$^\ddag$\\

\midrule

\multirow{5}{*}{\rotatebox[origin=c]{90}{Quantization}} &  $\boldsymbol{Q}\!=\!16$ & \it 10.64$^\dag$ & \it 13.58$^\dag$ & \it 26.34$^\dag$ & \it 21.15$^\dag$ & \it 29.9$^\ddag$\\
\cmidrule{2-7}
 & $\boldsymbol{Q}\!=\!8$ & \it 11.78$^\dag$ & \it 14.22$^\dag$ & \it 26.56$^\dag$ &  \it 21.63$^\dag$  & \it 30.10$^\ddag$\\
\cmidrule{2-7}
 
& $\boldsymbol{Q}\!=\!4$ & \it 83.89$^\dag$ & \it 36.64$^\dag$  & \it 35.89$^\dag$ &  \it 89.26$^\dag$ & \it 73.22$^\ddag$\\

\cmidrule{2-7}
 
& $\boldsymbol{Q}\!=\!2$ & \it 90.00$^\dag$ & \it 89.52$^\dag$ & \it 86.49$^\dag$ & \it 89.33$^\dag$ & \it 97.33$^\ddag$\\

\bottomrule
 \end{tabular}
}
\end{table*}

\subsection{Robustness evaluation}
\label{sec::rob}

Robustness evaluates the detector's capacity to retrieve the watermark of a watermarked content that has been altered. In Table~\ref{tab:robustness}, the watermarked neural network has been altered by the four attacks described in Sec.~\ref{sec::Testbed}. The columns correspond to the different setups (dataset and architecture) while the lines are labeled by the attack parameter and provide the performance metric of the task when the watermark has been retrieved. The Table is entirely filled since the watermark can be retrieved even if the performance of the watermarked model is very low (for instance, quantization for the values 2 and 4 bits).

\subsection{Security, data payload and computational cost }
\label{sec::secu}
%\vspace{-16pt}
\begin{table*}[h]
    \centering

\caption{Impact on the performance of removing the watermark using the secret key, depending on the number of watermarked weights. Values corresponds to the absolute variation of the complementary accuracy multiplied by 100.} 
\label{tab:keysteal}
\resizebox{.75\columnwidth}{!}{
\begin{tabular}{c c c c c c c c c c c c c c c  }
 \toprule
 &\multirow{2}{*}{\bf Method} &\multicolumn{6}{c}{$\bf |W_x| (\uparrow$)}\\
&  & $768$ & $3072$ & $12288$ & $49152$ & $196608$ & $786432$ \\
\midrule
 \multirow{5}{*}{\rotatebox[origin=c]{90}{VGG}} & Uchida~et~al.~\cite{uchida2017embedding} & \it +0 & \it +0 & \it -0.01 & $\star$ & $\star$ & $\star$ \\
& Tartaglione~et~al.~\cite{tartaglione2021delving} & \it +6.39 & \it +5.15 & \it +8.33 & \it +7.04 & $\star$ & $\star$ \\
& Li~et~al.~\cite{li2022fused} & \it +0 & \it +0 & \it +0 & \it +0 & \it +0 & \it +0 \\
 & Lv~et~al.~\cite{lv23watermarking} & \it +0 & \it +0 & \it +0 & \it +0 & \it +0 & \it +0 \\
& WaterMAS & \it +0.34 & \it +1.38 & \it +4.19 & \it +47.46 & $\star$ & $\star$ \\ 

\midrule

  \multirow{5}{*}{\rotatebox[origin=c]{90}{ResNet}} & Uchida~et~al.~\cite{uchida2017embedding} & \it + 0 & $\star$ & $\star$ & $\star$ & $\star$  & $\star$ \\
& Tartaglione~et~al.~\cite{tartaglione2021delving}& \it +29.75 & \it +24.93 & \it +26.58 & \it +22.15  & \it +25.94 & \it +25.96 \\
& Li~et~al.~\cite{li2022fused} & \it +0 & \it +0 & \it +0 & \it +0 & \it +0 & \it +0 \\
 & Lv~et~al.~\cite{lv23watermarking} & \it +0 & \it +0 & \it +0 & \it +0 & \it +0 & \it +0 \\
& WaterMAS&   \it +0.04 & \it +3.86 & \it +28.49 & \it +68.63 & \it +68.87 & \it +66.28 \\ 
\midrule

  \multirow{5}{*}{\rotatebox[origin=c]{90}{MobileNet}} & Uchida~et~al.~\cite{uchida2017embedding} & \it +0.04 & $\star$ & $\star$ & $\star$ & $\star$ & $\star$ \\
& Tartaglione~et~al.~\cite{tartaglione2021delving} & \it +44.57 & \it +34.47  & \it +38.58  & \it +37.51 & \it +34.47 &$\star$\\
  & Li~et~al.~\cite{li2022fused} & \it +0 & \it +0 & \it +0 & \it +0 & \it +0 & \it +0 \\
 & Lv~et~al.~\cite{lv23watermarking} & \it +0 & \it +0 & \it +0 & \it +0 & \it +0 & \it +0 \\
& WaterMAS&   \it +3.24 & \it +10.16 & \it +50.01 & \it +62.57 & \it +55.98 & $\star$\\ 

\midrule

  \multirow{5}{*}{\rotatebox[origin=c]{90}{SwinT}} & Uchida~et~al.~\cite{uchida2017embedding} & \it -0.03 & \it -0.17 & \it -0.56 & \it +0.04 & $\star$ & $\star$  \\
& Tartaglione~et~al.~\cite{tartaglione2021delving}&   \it +7.42 & \it +11.16 & \it +17.39 & \it +25.46 & \it +34.09 & $\star$\\
& Li~et~al.~\cite{li2022fused} & \it +0 & \it +0 & \it +0 & \it +0 & \it +0 & \it +0 \\
 & Lv~et~al.~\cite{lv23watermarking} & $\star$ & $\star$ & $\star$ & $\star$ & $\star$  & $\star$ \\
& WaterMAS&   \it +4.21 & \it +36.64  & \it +51.48 & \it +64.2 & \it +63.45 & $\star$\\ 

\bottomrule
 \end{tabular}
}
\end{table*}

Let's explore the number of watermarked weights that can convey the watermark. The results presented in Table~\ref{tab:keysteal}, show the absolute variation of the performance when altering the watermarking weights. The watermark is destroyed in all scenarios while $\star$ indicates that the watermark could not even be inserted. 
\cite{uchida2017embedding} was designed to fit the size of the inserted watermarked to the size of the layer. Even if this adaptation is possible, the watermark can be removed with a low impact on the performance. For \cite{li2022fused}, the identified neuron can be multiplied by $-1$ while its corresponding input channel of the next layer will also be multiplied by $-1$ resulting in an identic output value while the recovered watermark is destroyed. For~\cite{lv23watermarking}, the access to the HufuNet decoder allows the attacker to fine-tune the HufuNet model to avoid watermark detection without any modification on the watermarked model. For~\cite{tartaglione2021delving}, the performances of the model are lost when the watermark is removed. WaterMAS has a similar behavior as \cite{tartaglione2021delving}, but it also features an inverse linear dependency between the performance and the size of the removed watermark. This behavior opens the door for a larger data payload to be inserted by WaterMAS: while our experimental study does not explore this direction, the large span between the minimal and the maximal sizes can be exploited for increasing the data payload. 

%\vspace{-16pt}
\begin{table*}[h]
\centering
\caption{Watermark insertion computational cost: values extracted during the training, for the same batch size and architecture. MiB - mebibyte ($2^{20}$ Bytes).} \label{tab:compcost}
\resizebox{\textwidth}{!}{
\begin{tabular}{ c c c c c c c}
 \toprule
  \multirow{2}{*}{\bf Method}  & & \multicolumn{4}{c}{\bf CIFAR10} &\bf Cityscapes \\
    &  & VGG16 & ResNet18 & MobileNet  & SwinT & DeepLab-v3\\
\midrule
\multirow{2}{*}{Unwatermarked} &  GPU Memory  & 858 MiB & 502 MiB & 622 MiB & 1786 MiB & 6861 MiB \\
 & time$\slash$epoch & 12'' & 12'' & 13'' & 40'' & 4'20''\\
 \midrule
\multirow{2}{*}{Uchida~et~al.~\cite{uchida2017embedding}} & GPU Memory  & 858 MiB & 502 MiB & 622 MiB & 1786 MiB & 6877 MiB\\
 & time$\slash$epoch & 27'' & 49'' & 32'' & 1'14'' & 4'25''\\
 \midrule
\multirow{2}{*}{Tartaglione~et~al.~\cite{tartaglione2021delving}} & GPU Memory  & 4556 MiB & 592 MiB & 655 MiB & 2122 MiB & 7103 MiB\\
 & time$\slash$epoch & 1'33'' & 1'36'' & 8'32'' & 10'18'' & 16'10'' \\
  \midrule
\multirow{2}{*}{Lv~et~al.~\cite{lv23watermarking}} & GPU Memory  & 1855 MiB & 1354 MiB &  736 MiB & $\star$ & $\star$ \\
& time$\slash$epoch & 3'33'' & 4'06'' & 6'13'' & $\star$ &$ \star$\\
\midrule
\multirow{2}{*}{WaterMAS} & GPU Memory  & 1078 MiB & 656 MiB & 634 MiB & 2074 MiB & 7043 MiB\\
 & time$\slash$epoch & 27'' & 19'' & 57'' & 1'54'' & 8'44'' \\
\bottomrule
\end{tabular}
}
\end{table*}

In Table~\ref{tab:compcost}, the memory footprint and the mean time of execution per epoch are presented for the four methods \cite{uchida2017embedding,tartaglione2021delving,lv23watermarking}, the standard deviation does not appear since it was less than 1 second. \cite{li2022fused} does not appear in the table since there is no constraint on the training.
\section{Conclusion}
\label{sec:conclu}

With this paper, a new white-box watermarking method, referred to as WaterMAS, is advanced to reach the trade-off among robustness, imperceptibility, computational complexity, and data payload. First, by reconsidering and extending the SAM principles~\cite{foret2021sharpnessaware}, a new regularisation term is designed for sharpening the watermarked weights landscape. This way, the strength of the attacks that can be applied to WaterMAS is intrinsically reduced, and the robustness is ensured. Second, the imperceptibility property is reached by tightly coupling this regularization term with the training process. Third, the extra computational complexity required is one back-propagation step. Finally, the insertion occurs before training by randomly selecting a set of weights throughout the entire model, irrespective of the structures of layers. Compared to state-of-the-art methods~\cite{uchida2017embedding,li2022fused,lv23watermarking}, the main advantage is represented by the security, expressed as the possibility of keeping the watermark even when the secret key is intercepted, as presented in Table~\ref{tab:keysteal}. Compared to~\cite{tartaglione2021delving,lv23watermarking}, the main advantage is the computational cost reduction. Beyond watermarking, MAS opens the road to further explorations and applications of finding solutions in sharp loss minima, which can lead to sparse neural network representations~\cite{tartaglione2020pruning} or even further exploration of properties of these minima~\cite{dinh2017sharp}.

\bibliographystyle{splncs04}
\bibliography{main}

\end{document}